%% file: naaclhlt2018.tex
\documentclass[11pt,a4paper]{article}
\usepackage[hyperref]{naaclhlt2018}
\usepackage{times}
\usepackage{latexsym}

\usepackage{url}

\aclfinalcopy 
\def\aclpaperid{1102} 

\usepackage{amsmath}
\usepackage{amsfonts,amssymb}
\usepackage{color}
\usepackage{graphicx}
\graphicspath{ {figs/} }
\usepackage{float}
\usepackage{calc}
\usepackage{blindtext}
\usepackage{enumitem}
\usepackage{mwe}
\usepackage{booktabs}
\usepackage{multirow}
\usepackage{siunitx}

{{}}

\title{Object Ordering with Bidirectional Matchings for Visual Reasoning}

\author{Hao Tan \and Mohit Bansal \\
  UNC Chapel Hill \\
  {\tt \{haotan, mbansal\}@cs.unc.edu} \\
 }

\begin{document}

\maketitle

\begin{abstract} 
Visual reasoning with compositional natural language instructions, e.g., based on the newly-released Cornell Natural Language Visual Reasoning (NLVR) dataset, is a challenging task, where the model needs to have the ability to create an accurate mapping between the diverse phrases and the several objects placed in complex arrangements in the image. Further, this mapping needs to be processed to answer the question in the statement given the ordering and relationship of the objects across three similar images. In this paper, we propose a novel end-to-end neural model for the NLVR task, where we first use joint bidirectional attention to build a two-way conditioning between the visual information and the language phrases. Next, we use an RL-based pointer network to sort and process the varying number of unordered objects (so as to match the order of the statement phrases) in each of the three images and then pool over the three decisions. Our model achieves strong improvements (of 4-6\% absolute) over the state-of-the-art on both the structured representation and raw image versions of the dataset.
\end{abstract}

\input{1_intro}
\input{2_related}

\input{3_model}
\input{4_experiments}
\input{5_results}

\vspace{-4pt}
\section{Conclusion}
\vspace{-4pt}
We presented a novel end-to-end model with joint bidirectional attention and object-ordering pointer networks for visual reasoning. We evaluate our model on both the structured-representation and raw-image versions of the NLVR dataset and achieve substantial improvements over the previous end-to-end state-of-the-art results.

\vspace{-5pt}
\section*{Acknowledgments}
\vspace{-5pt}
We thank the anonymous reviewers for their helpful comments. This work was supported by a
Google Faculty Research Award, a Bloomberg Data Science Research Grant, an IBM Faculty
Award, and NVidia GPU awards.

\bibliography{naaclhlt2018}
\bibliographystyle{acl_natbib}

\appendix
\input{6_supp.tex}
\end{document}

%% file: 1_intro.tex
\section{Introduction}
\begin{figure}
\centering
\includegraphics[width=0.48\textwidth]{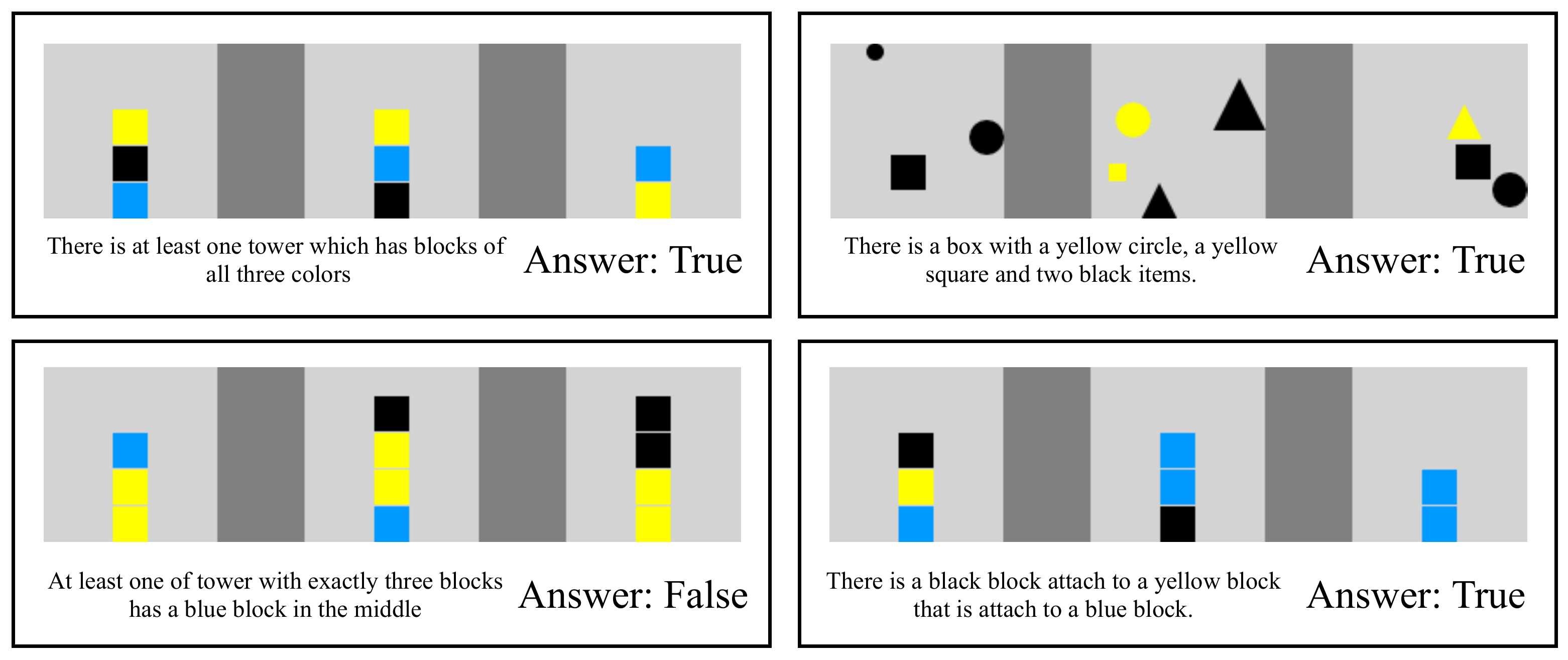}
\vspace{-21pt}
        \caption{NLVR task: given an image with 3 sub-images and a statement, the model needs to predict whether the statement correctly describes the image or not. We show 4 such examples which our final BiATT-Pointer model correctly classifies but the strong baseline models do not (see Sec.~\ref{sec:results}).
        \vspace{-15pt}
}
        \label{fig:task}
\end{figure}

Visual Reasoning \cite{antol2015vqa,  andreas2016neural, bisk2016natural, johnson2017clevr} requires a sophisticated understanding of the compositional language instruction and its relationship with the corresponding image. \newcite{suhr2017corpus} recently proposed a challenging new NLVR task and dataset in this direction with natural and complex language statements that have to be classified as true or false given a multi-image set (shown in Fig.~\ref{fig:task}). Specifically, each task instance consists of an image with three sub-images and a statement which describes the image. The model is asked to answer the question whether the given statement is consistent with the image or not.

To solve the task, the designed model needs to fuse the information from two different domains, the visual objects and the language, and learn accurate relationships between the two. Another difficulty is that the objects in the image do not have a fixed order and the number of objects also varies. Moreover, each statement reasons for truth over three sub-images (instead of the usual single image setup), which also breaks most of the existing models. In our paper, we introduce a novel end-to-end model to address these three problems, leading to strong gains over the previous best model. Our pointer network based LSTM-RNN sorts and learns recurrent representations of the objects in each sub-image, so as to match it better with the order of the phrases in the language statement. For this, it employs an RL-based policy gradient method with a reward extracted from the subsequent comprehension model. With these strong representations of the visual objects and the statement units, a joint-bidirectional attention flow model builds consistent, two-way matchings between the representations in different domains. 
Finally, since the scores computed by the bidirectional attention are about the three sub-images, a pooling combination layer over the three sub-image representations is required to give the final score of the whole image.

On the structured-object-representation version of the dataset, our pointer-based, end-to-end bidirectional attention model achieves an accuracy of 73.9\%, outperforming the previous (end-to-end) state-of-the-art method by 6.2\% absolute, where both the pointer network and the bidirectional attention modules contribute significantly. We also contribute several other strong baselines for this new NLVR task based on Relation Networks \cite{santoro2017simple} and BiDAF \cite{seo2016bidirectional}. 
Furthermore, we also show the result of our joint bidirectional attention model on the raw-image version (with pixel-level, spatial-filter CNNs) of the NLVR dataset, where our model achieves an accuracy of 69.7\% and outperforms the previous best result by 3.6\%. On the unreleased leaderboard test set, our model achieves an accuracy of 71.8\% and 66.1\% on the structured and raw-image versions, respectively, leading to 4\% absolute improvements on both tasks. Finally, we present analysis of the pointer network's learned object order as well as success and failure examples of the overall model.

%% file: 2_related.tex
\begin{figure*}
\includegraphics[width=\textwidth]{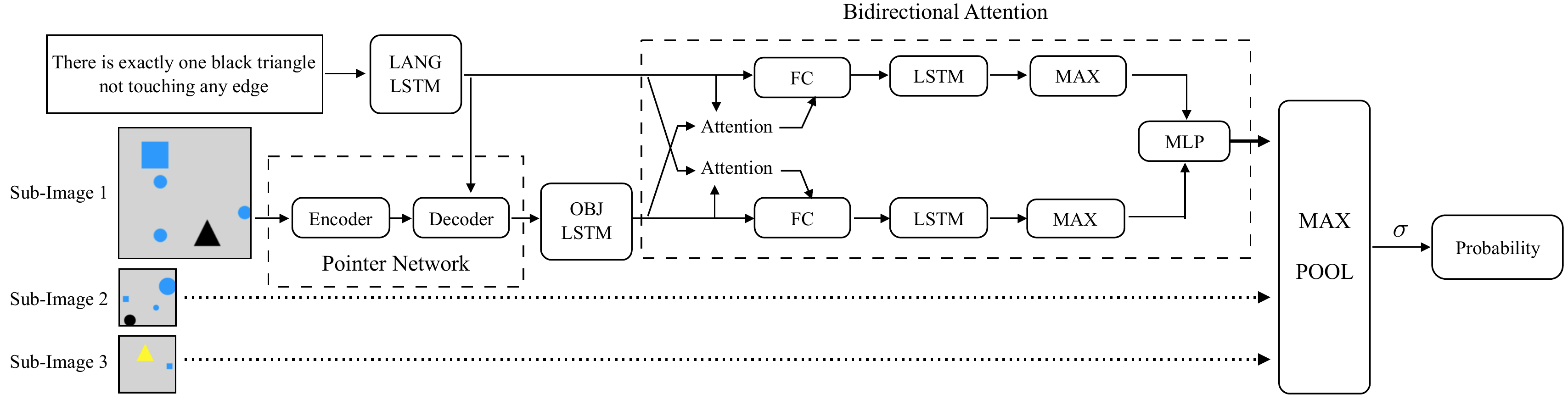}
\vspace{-25pt}
        \caption{Our BiATT-Pointer model with a pointer network and a joint bidirectional attention module.}
        \label{fig:model}
        \vspace{-10pt}
\end{figure*}

\vspace{-2pt}
\section{Related work}
\vspace{-3pt}
Besides the NLVR corpus with a focus on complex and natural compositional language~\cite{suhr2017corpus}, other useful visual reasoning datasets have been proposed for navigation and assembly tasks~\cite{macmahon2006walk, bisk2016natural}, as well as for visual Q\&A tasks which focus more on complex real-world images~\cite{antol2015vqa,johnson2017clevr}.
Specifically for the NLVR dataset, previous models have incorporated property- and count-based features of the objects and the language~\cite{suhr2017corpus}, or extra semantic parsing (logical form) annotations~\cite{goldman2017weakly} -- we focus on end-to-end models for this visual reasoning task.

Attention mechanism~\cite{bahdanau2014neural, luong2015effective,xu2015show} has been widely used for conditioned language generation tasks.
It is further used to learn alignments between different modalities~\cite{lu2016hierarchical,wang2016machine,seo2016bidirectional,andreas2016neural,chaplot2017gated}. 
In our work, a bidirectional attention mechanism is used to learn a joint representation of the visual objects and the words by building matchings between them.
 
Pointer network~\cite{vinyals2015pointer} was introduced to learn the conditional probability of an output sequence. \newcite{bello2016neural} extended this to near-optimal combinatorial optimization via reinforcement learning.
In our work, a policy gradient based pointer network is used to ``sort'' the objects conditioned on the statement, such that the sequence of ordered objects is sent to the subsequent comprehension model for a reward.

%% file: 3_model.tex
\section{Model}
The training datum for this task consists of the statement $s$, the structured-representation objects $o$ in the image $I$, and the ground truth label $y$ (which is $1$ for true and $0$ for false).
Our BiATT-Pointer model (shown in Fig.~\ref{fig:model}) for the structured-representation task uses the pointer network to sort the object sequence (optimized by policy gradient), and then uses the comprehension model to calculate the probability $P(s, o)$ of the statement $s$ being consistent with the image.
Our CNN-BiATT model for the raw-image $I$ dataset version is similar but learns the structure directly via pixel-level, spatial-filter CNNs -- details in Sec.~\ref{sec:results} and the appendix.
In the remainder of this section, we first describe our BiATT comprehension model and then the pointer network. 

\subsection{Comprehension Model with Joint Bidirectional Attention}
\label{sec:cph}

We use one bidirectional LSTM-RNN~\cite{hochreiter1997long}
(denoted by LANG-LSTM) to read the statement $s=w_1, w_2, \ldots, w_\textsc{T}$, and output the hidden state representations $\{h_i\}$.
A word embedding layer is added before the LSTM to project the words to high-dimension vectors $\{\tilde{w}_i\}$.
\vspace{-5pt}
\begin{align}
h_1, & h_2  ,  \dotsc, h_\textsc{T}  = \mathrm{LSTM} \left(\tilde{w}_1, \tilde{w}_2, \ldots	, \tilde{w}_\textsc{T} \right)
\label{eqn:langlstm}
\vspace{-5pt}
\end{align}
The raw features of the objects in the $j$-th sub-image are $\{o_k^j\}$ (since the NLVR dataset has 3 sub-images per task).
A fully-connected (FC) layer without nonlinearity projects the raw features to object embeddings $\{e_k^j\}$.
We then go through all the objects in random order (or some learnable order, e.g., via our pointer network, see Sec.~\ref{sec:pointer_net}) by another bidirectional LSTM-RNN (denoted by OBJ-LSTM), whose output is a sequence of vectors $\{g_k^j\}$ which is used as the (left plus right memory) representation of the objects (the objects in different sub-images are handled separately):
\vspace{-5pt}
\begin{align}
e_k^j &= W \, o_k^j +  b \\  
g_1^j, g_2^j, \ldots, g_{\textsc{N}_j}^j  &= \mathrm{LSTM}\, ( e_1^j, e_2^j, \ldots, e_{\textsc{N}_j}^j ) 
\end{align}
where $\textsc{N}_j$ is the number of the objects in $j$th sub-image.
Now, we have two vector sequences for the representations of the words and the objects, using which the bidirectional attention then calculates the score measuring the correspondence between the statement and the image's object structure.
To simplify the notation, we will ignore the sub-image index $j$. We first merge the LANG-LSTM hidden outputs $\{h_i\}$ and the object-aware context vectors $\{c_i\}$ together to get the joint representation $\{\hat{h}_i\}$. The object-aware context vector $c_i$ for a particular word $w_i$ is calculated based on the bilinear attention between the word representation $h_i$ and the representations of the objects $\{g_k\}$: 
\begin{align}
\vspace{-1pt}
\alpha_{i,k} &= \mathrm{softmax}_k \left( h_i^\intercal \, B_\textsc{1} \, g_k \right) \\
c_i & = \sum_k \alpha_{i, k} \cdot g_k \\
\hat{h}_i & = \mathrm{relu} \left( W_\textsc{LANG} \left[ h_i; \, c_i ; \, h_i \mbox{-}  c_i; \, h_i \! \circ \! c_i \right] \right)
\vspace{-1pt}
\end{align}
where the symbol $\circ$ denotes element-wise multiplication.

\paragraph{Improvement over BiDAF} The BiDAF model of~\newcite{seo2016bidirectional} does not use a full object-to-words attention mechanism. The query-to-document attention module in BiDAF added the attended-context vector to the document representation instead of the query representation. However, the inverse attention from the objects to the words is important in our task because the representation of the object depends on its corresponding words. 
Therefore, different from the BiDAF model, we create an additional `symmetric' attention to merge the OBJ-LSTM hidden outputs $\{g_k\}$ and the statement-aware context vectors $\{d_k\}$  together to get the joint representation $\{\hat{g}_k\}$. The improvement (6.1\%) of our BiATT model over the BiDAF model is shown in Table~\ref{table:final_result}. 
\begin{align}
  \beta_{k,i} &= \mathrm{softmax}_i \left( g_k^\intercal \, B_\textsc{2} \, h_i \right) \\
d_k & = \sum_i \beta_{k, i} \cdot h_i \\
\hat{g}_k & = \mathrm{relu} \left( W_\textsc{OBJ} \left[ g_k;  d_k ;  g_k\mbox{-}d_k;  g_k \! \circ \! d_k \right] \right)
\end{align}
These above vectors $\{\hat{h}_i\}$ and $\{\hat{g}_k\}$ are the representations of the words and the objects which are aware of each other bidirectionally. To make the final decision, two additional bidirectional LSTM-RNNs are used to further process the above attention-based representations via an additional memory-based layer. Lastly, two max pooling layers over the hidden output states create two single-vector outputs for the statement and the sub-image, respectively:
\begin{align}
\bar{h}_1, \bar{h}_2, \ldots, \bar{h}_\textsc{T} & = \mathrm{LSTM} (\hat{h}_1, \hat{h}_2, \ldots, \hat{h}_\textsc{T} ) \\
\bar{g}_1, \bar{g}_2, \ldots, \bar{g}_\textsc{N} & = \mathrm{LSTM} (\hat{g}_1, \hat{g}_2, \ldots, \hat{g}_\textsc{N} ) \\
\bar{h} &= \mathrm{ele} \max_i  \left\{ \bar{h}_i\right\} \\
\bar{g} &= \mathrm{ele} \max_k \left\{ \bar{g}_k \right\}
\end{align}
where the operator $\mathrm{ele} \max$ denotes the element-wise maximum over the vectors.
The final scalar score for the sub-image is given by a 2-layer MLP over the concatenation of $\bar{h}$ and $\bar{g}$ as follows:
\begin{equation}
\mathit{score} = W_2 \, \tanh \left( W_1 [\bar{h}; \bar{g} ] + b_1 \right) 
\label{eqn:score}
\end{equation}

\paragraph{Max-Pooling over Sub-Images}
In order to address the 3 sub-images present in each NLVR task, a max-pooling layer is used to combine the above-defined scores of the sub-images.
Given that the sub-images do not have any specific ordering among them (based on the data collection procedure~\cite{suhr2017corpus}), a pooling layer is suitable because it is permutation invariant.
Moreover, many of the statements are about the existence of a special object or relationship in one sub-image (see Fig.~\ref{fig:task}) and hence the max-pooling layer effectively captures the meaning of these statements.
We also tried other combination methods (mean-pooling, concatenation, LSTM, early pooling on the features/vectors, etc.); the max pooling (on scores) approach was the simplest and most effective method among these (based on the dev set).

The overall probability that the statement correctly describes the full image (with three sub-images) is the sigmoid of the final max-pooled score. The loss of the comprehension model is the negative log probability (i.e., the cross entropy):
\begin{align}
P(s, o) =& \, \sigma \left( \max_j    \mathit{score}^j \right) \\
L(s, o, y) =& - y \, \log \! P(s, o) \nonumber \\
\,&- (1-y)\,  \log (1 - P(s,o)) 
\label{eqn:loss}
\end{align}
where $y$ is the ground truth label.

\subsection{Pointer Network}
\label{sec:pointer_net}
Instead of randomly ordering the objects, humans look at the objects in an appropriate order w.r.t. their reading of the given statement and after the first glance of the image.
Following this idea, we use an additional pointer network~\cite{vinyals2015pointer} to find the best object ordering for the subsequent language comprehension model. The pointer network contains two RNNs, the encoder and the decoder. 
The encoder reads all the objects in a random order.
The decoder then learns a permutation $\pi$ of the objects' indices, by recurrently outputting a distribution over the objects based on the attention over the encoder hidden outputs. At each time step, an object is sampled without replacement following this distribution. Thus, the pointer network models a distribution $p(\pi \mid s, o)$ over all the permutations:
\begin{equation}
p(\pi \mid s, o) = \prod_i p\left(\pi(i) \mid \pi(<i), s, o\right)
\end{equation}
Furthermore, the appropriate order of the objects depends on the language statement, and hence the decoder importantly attends to the hidden outputs of the LANG-LSTM (see Eqn.~\ref{eqn:langlstm}).

The pointer network is trained via reinforcement learning (RL) based policy gradient optimization.
The RL loss $L_\textsc{RL}(s,o,y)$ is defined as the expected comprehension loss (expectation over the distribution of permutations):
\begin{equation}
L_\textsc{RL}  (s,o,y)  = E_{\pi \sim p(\cdot
\mid s, o)} L(s, o[\pi], y)  
\end{equation}
where $o[\pi]$ denotes the permuted input objects for permutation $\pi$, and $L$ is the loss function defined in Eqn.~\ref{eqn:loss}.
Suppose that we sampled a permutation $\pi^*$ from the distribution $p(\pi|s,o)$; then the above RL loss could be optimized via policy gradient methods~\cite{williams1992simple}.  The reward $R$ is the negative loss of the subsequent comprehension model $L(s,o[\pi^*],y)$.  A baseline $b$ is subtracted from the reward to reduce the variance (we use the self-critical baseline of~\newcite{rennie2016self}). The gradient of the loss $L_\textsc{RL}$ could then be approximated as:
\begin{align}
R  = &- L(s, o[\pi^*], y) \\
\nabla_\theta L_\textsc{RL}(s,o,y) \approx & \, -  (R - b) \nabla_\theta \log p(\pi^* \mid s, o) \nonumber \\
&\, + \nabla_\theta L(s,o[\pi^*],y)
\end{align}
This overall BiATT-Pointer model (for the structured-representation task) is shown in Fig.~\ref{fig:model}.

%% file: 4_experiments.tex
\section{Experimental Setup}
We evaluate our model on the NLVR dataset~\cite{suhr2017corpus}, for both the structured and raw-image versions. 
All model tuning was performed on the dev set. Given the fact that the dataset is balanced (the number of true labels and false labels are roughly the same), the accuracy of the whole corpus is used as the metric.
We only use the raw features of the statement and the objects with minimal standard preprocessing (e.g., tokenization and UNK replacement; see appendix for reproducibility training details).

%% file: 5_results.tex
\begin{table*}[t]
\begin{center}
\begin{small}
\begin{tabular}{|p{6cm}|p{2cm}p{2cm}c|} 
\hline
Model & Dev & Test-P & Test-U \\
\hline\hline
\multicolumn{4}{|c|}{ STRUCTURED REPRESENTATIONS DATASET }\\
\hline\hline
MAXENT \cite{suhr2017corpus} & 68.0\% & 67.7\% & 67.8\% \\
MLP \cite{suhr2017corpus} & 67.5\% & 66.3\% & 65.3\% \\
ImageFeat+RNN \cite{suhr2017corpus} & 57.7\% & 57.6\% & 56.3\% \\ 
\hline
RelationNet~\cite{santoro2017simple}    & 65.1\% & 62.7\% & - \\
 BiDAF \cite{seo2016bidirectional} & 66.5\% & 68.4\% & -  \\
\hline
BiENC Model & 65.1\% & 63.4\% & -  \\
BiATT Model & 72.6\% & 72.3\% & - \\
BiATT-Pointer Model & \textbf{74.6\%} & \textbf{73.9\%}  & \textbf{71.8\%} \\
\hline\hline
\multicolumn{4}{|c|}{RAW IMAGE DATASET}\\
\hline\hline
CNN+RNN \cite{suhr2017corpus} & 56.6\% & 58.0\% & 56.3\% \\
NMN \cite{suhr2017corpus} & 63.1\% & 66.1\% & 62.0\% \\
\hline
CNN-BiENC Model & 58.7\% & 58.7\%  & - \\
CNN-BiATT Model & \textbf{66.9\%} & \textbf{69.7\%} & \textbf{66.1\%} \\
\hline
\end{tabular}
\end{small}
\end{center}
\vspace{-10pt}
\caption{Dev, Test-P (public), and Test-U (unreleased) results of our model on the structured-representation and raw-image datasets, compared to the previous SotA results and other reimplemented baselines. 
}
\vspace{-8pt}
\label{table:final_result}
\end{table*}

\section{Results and Analysis}
\label{sec:results}

\noindent\textbf{Results on Structured Representations Dataset}: Table~\ref{table:final_result} shows our primary model results. In terms of previous work, the state-of-the-art result for end-to-end models is `MAXENT', shown in \newcite{suhr2017corpus}.\footnote{There is also recent work by~\newcite{goldman2017weakly}, who use extra, manually-labeled semantic parsing data to achieve a released/unreleased test accuracy of 80.4\%/83.5\%, resp.}
Our proposed BiATT-Pointer model (Fig.~\ref{fig:model}) achieves a 6.2\% improvement on the public test set and a 4.0\% improvement on the unreleased test set over this SotA model.
To show the individual effectiveness of our BiATT and Pointer components, we also provide two ablation results:
(1) the bidirectional attention BiATT model without the pointer network; and (2) our BiENC baseline model without any attention or the pointer mechanisms. 
The BiENC model uses the similarity between the last hidden outputs of the LANG-LSTM and the OBJ-LSTM as the score (Eqn.~\ref{eqn:score}).

Finally, we also reproduce some recent popular frameworks, i.e., Relationship Network~\cite{santoro2017simple} and BiDAF model~\cite{seo2016bidirectional}, which have been proven to be successful in other machine comprehension and visual reasoning tasks. 
The results of these models are weaker than our proposed model. Reimplementation details are shown in the appendix.

\noindent\textbf{Results on Raw Images Dataset}: To further show the effectiveness of our BiATT model, we apply this model to the raw image version of the NLVR dataset, with minimal modification. We simply replace each object-related LSTM with a visual feature CNN that directly learns the structure via pixel-level, spatial filters (instead of a pointer network which addresses an unordered sequence of structured object representations). As shown in Table ~\ref{table:final_result}, this CNN-BiATT model outperforms the neural module networks (NMN) \cite{andreas2016neural} previous-best result by 3.6\% on the public test set and 4.1\% on the unreleased test set. More details and the model figure are in the appendix.

\begin{figure}
\centering
\includegraphics[width=0.48\textwidth]{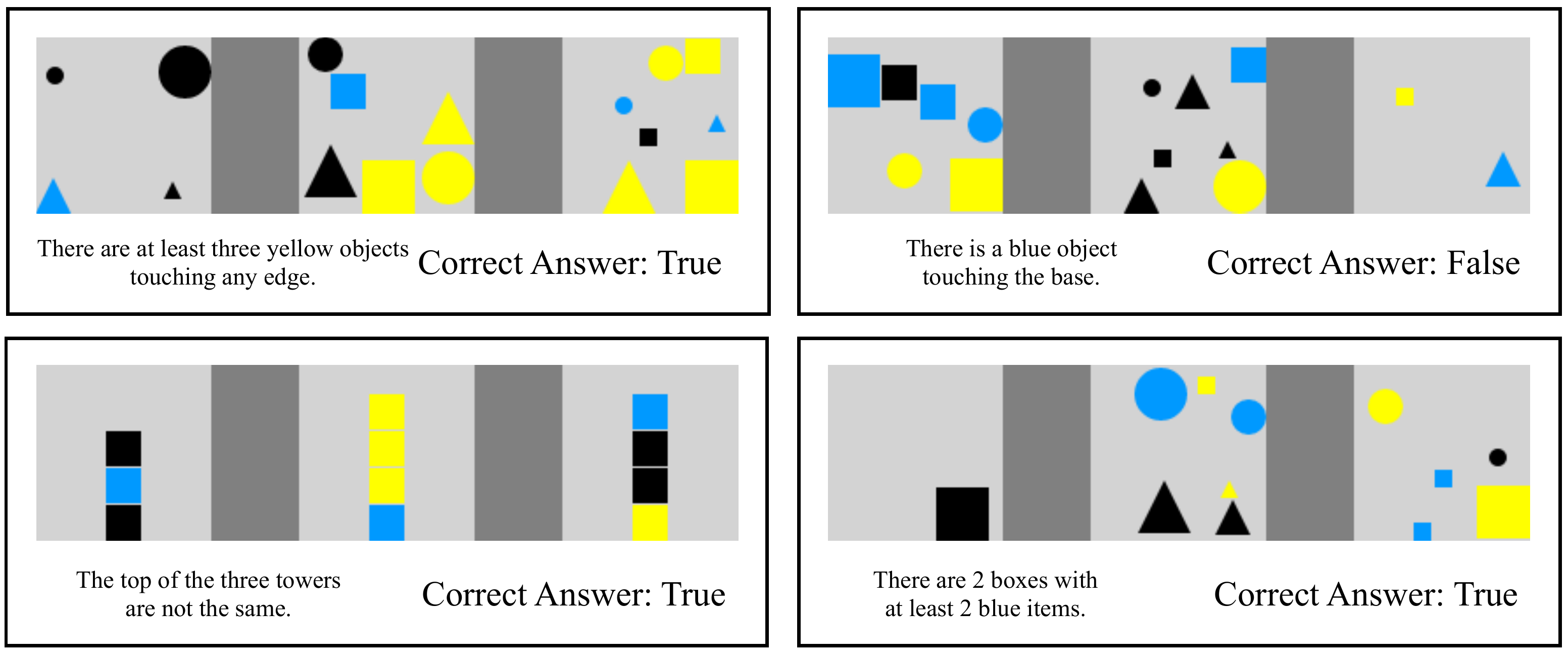}
\vspace{-25pt}
        \caption{Incorrectly-classified examples.}
        \label{fig:neg_example}
        \vspace{-8pt}
\end{figure}

\begin{figure}
\centering
\includegraphics[width=0.48\textwidth]{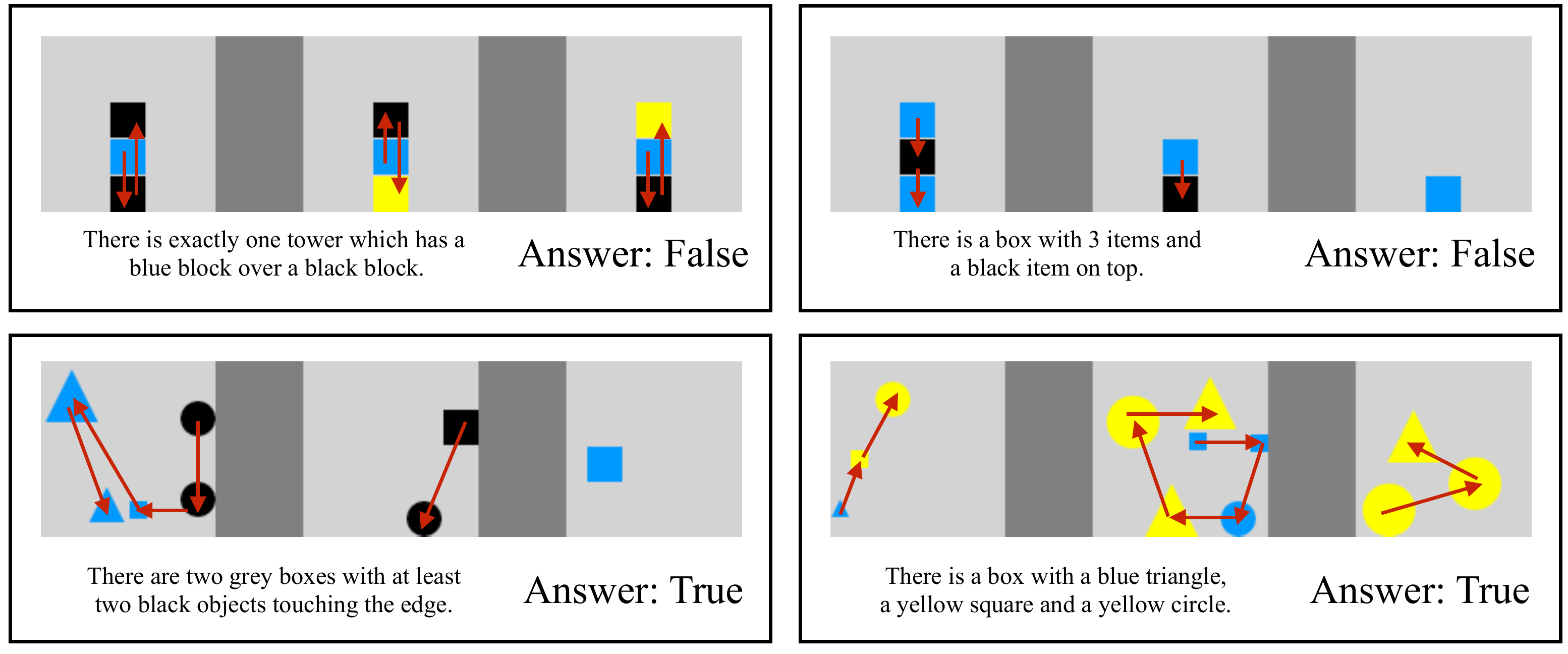}
\vspace{-25pt}
        \caption{Examples of our learned object ordering. The red arrows indicate the order of the objects learned by the pointer network.}
        
        \label{fig:pointer}
        \vspace{-10pt}
\end{figure}

\noindent\textbf{Output Example Analysis}: Finally, in Fig.~\ref{fig:task}, we show some output examples which were successfully solved by our BiATT-Pointer model but failed in our strong baselines. The left two examples in Fig.~\ref{fig:task} could not be handled by the BiENC model. The right two examples are incorrect for the BiATT model without the ordering-based pointer network.
Our model can quite successfully understand the complex meanings of the attributes and their relationships with the diverse objects, as well as count the occurrence of and reason over objects without any specialized features. 

Next, in Fig.~\ref{fig:neg_example}, we also show some negative examples on which our model fails to predict the correct answer. 
The top two examples involve complex high-level phrases e.g., ``touching any edge'' or ``touching the base'', which are hard for an end-to-end model to capture, given that such statements are rare in the training data.
Based on the result of the validation set, the max-pooling layer is selected as the combination method in our model. 
The max-pooling layer will choose the highest score from the sub-images as the final score. Thus, the layer could easily handle statements about single-subimage-existence based reasoning (e.g., the 4 positively-classified examples in Fig.~\ref{fig:task}). 
However, the bottom two negatively-classified examples in Fig.~\ref{fig:neg_example} could not be resolved because of the limitation of the max-pooling layer on scenarios that consider multiple-subimage-existence.
We did try multiple other pooling and combination methods, as mentioned in Sec.~\ref{sec:cph}. Among these methods, the concatenation, early pooling and LSTM-fusion approaches might have the ability to solve these particular bottom-two failed statements. In our future work, we are addressing multiple types of pooling methods jointly. 

Finally, we show the effectiveness of the pointer network in learning the object order, in Fig.~\ref{fig:pointer}. The red arrows indicate the sorted order of the objects as learned by our pointer network conditioned on the language instruction. In the top two examples, the model learns to sort the objects in a path which is in accordance with the spatial relationships in the statement (e.g., ``blue block over a black block'' or ``item on top''). In the bottom two examples, the model also tries to learn the order of the objects that is aligned well with the occurrences of the words in the statement.

%% file: 6_supp.tex
\section{Supplementary Material}

\begin{figure*}
\includegraphics[width=\textwidth]{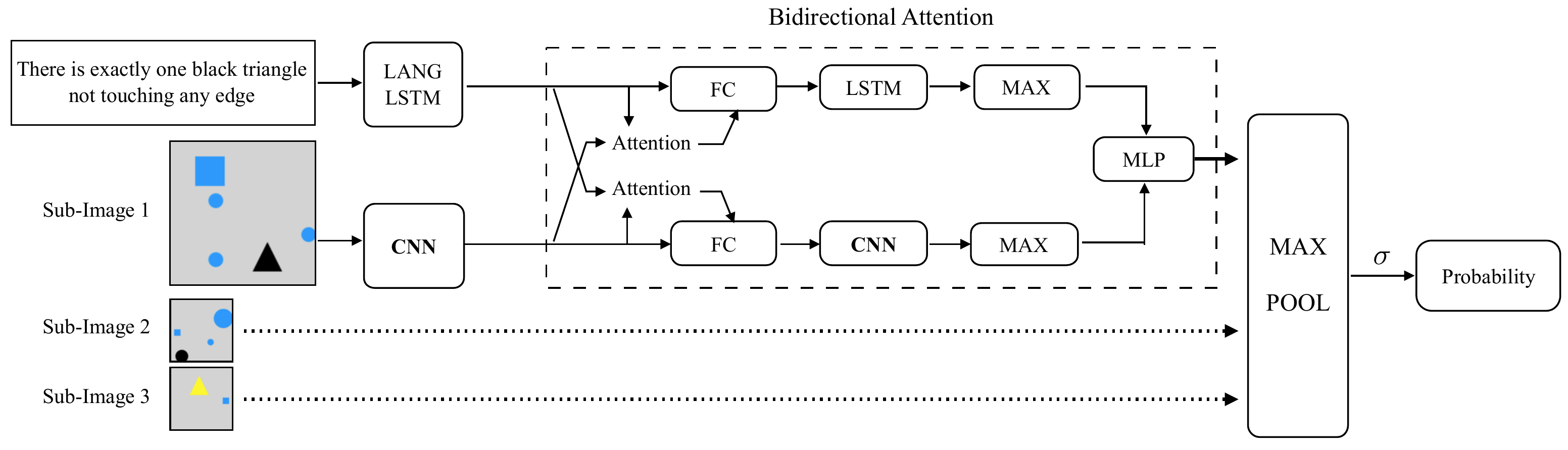}
\vspace{-20pt}
        \caption{Our CNN-BiATT model for the raw-image dataset version replaces every object-related LSTM-RNN with a spatial-filter convolutional neural network (CNN).  The CNN for the raw image-pixels is a pretrained ResNet-v2-101. A 3-layers CNN with relu activation is used in the bidirectional attention.}
        \label{fig:image_model}
\end{figure*}

\subsection{CNN-BiATT Model Details}
\label{sec:image_model}
As shown in Fig.~\ref{fig:image_model}, we apply our BiATT model to the raw image dataset with minimal modification. 
The visual input of the model for this task is changed from the unordered structured representation set of objects $o$ to the raw image pixels $I$.
Hence, we replace all object-related LSTMs (e.g., the OBJ-LSTM and the LSTM-RNN in the bidirectional attention in Fig.~\ref{fig:model}) with visual feature convolutional neural networks (CNNs) that directly learn the structure via pixel-level, spatial filters (instead of a pointer network which addresses an unordered sequence of structured object representations).

The training datum for the NLVR raw-image version consists of the statement $s$, the image $I$ and the ground truth label $y$. The image $I$ contains three sub-images $x^1$, $x^2$ and $x^3$. We will use $x$ to indicate any sub-image. The superscript which indicates the index of the sub-image is ignored to simplify the notation.
The representation of the statement $\{h_i\}$ is calculated by the LANG-LSTM as before. 
For the image representation, we project the sub-image to a sequence of feature vectors (i.e., the feature map) $\{a_l : l = 1,\ldots, L\}$ corresponding to 
the different image locations. $L=m\times m$ is the size of the features and $m$ is the width and height of the feature map. The projection consists of ResNet-V2-101~\cite{he2016identity} and a following fully-connected (FC) layer. We only use the blocks in the ResNet before the average pooling layer and the output of the ResNet is a feature map of size $m \! \times \! m \! \times \! 2048$.
\begin{align}
f_1, \ldots, f_{L} &= \mathrm{ResNet} (x) \\
 a_l & = \mathrm{relu} (W_{x}\,  f_l + b_x)
\end{align}
The joint-representation of the statement $\{\hat{h}_i \}$ is the combination of the LANG-LSTM hidden output states $\{h_i\}$ and the image-aware context vectors $\{c_i\}$:
\begin{align}
\alpha_{i,l} &= \mathrm{softmax}_l \left( h_i^\intercal \, B_\textsc{1} \, a_l \right) \\
c_i & = \sum_l \alpha_{i, l} \cdot a_l \\
\hat{h}_i & = \mathrm{relu} \left( W_\textsc{LANG} \left[ h_i; \, c_i ; \, h_i  \scalebox{0.9}[1.0]{\( - \)}  c_i; \, h_i \! \circ \! c_i \right] \right)
\end{align}
The joint-representation of the image $\{\hat{a}_l\}$ is calculated in the same way:
\begin{align}
  \beta_{l,i} &= \mathrm{softmax}_i \left( a_l^\intercal \, B_\textsc{2} \, h_i \right) \\
d_l & = \sum_i \beta_{l, i} \cdot h_i \\
\hat{a}_l & = \mathrm{relu} \left( W_\textsc{IMG} \left[ a_l;  d_l ;  a_l  \scalebox{0.9}[1.0]{\( - \)}  d_l;  a_l \! \circ \! d_l \right] \right)
\end{align}
The joint-representation of the statement is further processed by a LSTM-RNN. Different from our BiATT model, a 3-layers CNN is used for modeling the joint-representation of the image $\{\hat{a}_l\}$. The output of the CNN layer is another feature map $\{\bar{a}_l\}$.
Each CNN layer has kernel size $3 \times 3$ and uses relu as the activation function, and then we finally use element-wise max operator similar to Sec.~\ref{sec:cph}:
\begin{align}
\bar{h}_1, \bar{h}_2, \ldots, \bar{h}_\textsc{T} & = \mathrm{LSTM} (\hat{h}_1, \hat{h}_2, \ldots, \hat{h}_\textsc{T} ) \\
\bar{a}_1, \bar{a}_2, \ldots, \bar{a}_\textsc{L'} & = \mathrm{CNN} (\hat{a}_1, \hat{a}_2, \ldots, \hat{a}_\textsc{L} ) \\
\bar{h} &= \mathrm{ele} \max_i  \left\{ \bar{h}_i\right\} \\
\bar{a} &= \mathrm{ele} \max_l \left\{ \bar{a}_l \right\}
\end{align}
At last, we use the same method as our BiATT model to calculate the score and the loss function:
\begin{align}
\mathit{score} (s, x) = &W_2 \, \tanh \left( W_1 [\bar{h}; \bar{a} ] + b_1 \right)  \\
P(s, I) =& \, \sigma \left( \max_j \,   \mathit{score} (s, x^j) \right)  \\
L(s, I, y) =& - y \, \log \! P(s, I) \nonumber \\
\,&- (1-y)\,  \log (1 - P(s,I)) 
\end{align}

\subsection{Reimplementation Details for Relationship Network and BiDAF Models}
\label{sec:reimp}
We reimplement a Relationship Network~\cite{santoro2017simple}, using a three-layer MLP with 256 units per layer in the G-net and a three-layer MLP consisting of 256, 256 (with 0.3 dropout), and 1 units with ReLU nonlinearities for F-net. We also reimplement a BiDAF model~\cite{seo2016bidirectional} using 128-dimensional word embedding, 256-dimensional LSTM-RNN and 0.3 dropout rate. A max pooling layer on top of the modeling layer of BiDAF is used to merge the hidden outputs to a single vector. 

\subsection{Experimental Setup and Training Details for Our BiATT-Pointer, BiENC, and CNN-BiATT Models}
\label{sec:supp_setup}
\subsubsection{BiATT-Pointer}
For preprocessing, we replace the words whose occurrence is less than $3$ with the ``UNK'' token. 
We create a $9$ dimension vector as the feature of each object. This feature contains the location $(x,y)$ in 2D coordinate, the size of the object and two 3-dimensional hot vectors for the shape and the color. The $(x,y)$ coordinates are normalized to the range $[-1, 1]$.

For the model hyperparameters (all lightly tuned on dev set),  the dimension of the word embedding is 128, and the number of units in an LSTM cell is 256. The word embedding is trained from scratch. 
The object feature is projected to a 64-dimensional vector. 
The dimensions of joint representation $\hat{h}_i$ and $\hat{g}_k$ are both 512. The first fully-connected layer in calculating the sub-images score has 512 units. All the trainable variables are initialized with the Xavier initializer. To regularize the training process, we add a dropout rate $0.3$ to the hidden output of the LSTM-RNNs and before the last MLP layer which calculates the score for sub-images. We also clip the gradients by their norm to avoid gradient exploding. The losses are optimized by a single Adam optimizer and the learning rate is fixed at 1e-4. 

For the pointer network, we sample the objects following the distribution of the objects at each decoder step during training. In inference, we select the object with maximum probability. We use the self-critical baseline~\cite{rennie2016self} to stabilize the RL training, where the final score in inference (choosing object with maximum probability) is subtracted from the reward. To reduce the number of parameters, we share the weight of the fully-connected layer which projects the raw object feature to the high dimensional vector in the pointer encoder, the pointer decoder, and the OBJ-LSTM. The pointer decoder attends to the hidden outputs of the LANG-LSTM using bilinear attention ~\cite{luong2015effective}.

\subsubsection{CNN-BiATT}
We initialize our model with weights of the public pretrained ResNet-V2-101 (based on the ImageNet dataset) and freeze it during training. The ResNet projects the sub-image to a feature map of $10\! \times \! 10 \times \! 2048$. The feature map is normalized to a mean of 0 and a standard deviation of 1 before feeding into the FC layer. 
The fully connected layer after the ResNet has 512 units. 
Each layer of the 3-layers CNN in the bidirectional attention has kernel size $3\times 3$ with $512$ filters and no padding. 

\subsubsection{BiENC}
The BiENC model uses LANG-LSTM and OBJ-LSTM to read the statement and the objects. A bilinear form calculates the similarity between the last hidden outputs of the two LSTM-RNNs. The similarity is directly used as the score of the sub-image. The CNN-BiENC model replaces the OBJ-LSTM with a CNN.